\documentclass[11pt]{article}

\usepackage{acl}

\usepackage{times}
\usepackage{latexsym}
\usepackage[T1]{fontenc}
\usepackage[utf8]{inputenc}
\usepackage{microtype}

\usepackage{amsmath}
\usepackage{amssymb}
\usepackage{bm}

\usepackage{graphicx}
\usepackage{booktabs}
\usepackage{array}
\usepackage{multirow}

\usepackage{url}

\title{Bias-Corrected Ceilings of Emotion Predictability from Human Label
Variation Based on Instance-Level Fano Bounds}

\author{Keito Inoshita \\
  Faculty of Business and Commerce, Kansai University \\
  \texttt{inosita.2865@gmail.com}}

\begin{document}
\maketitle

\begin{abstract}
Emotion recognition from text keeps improving on benchmarks, yet whether an
accuracy ceiling has been reached is seldom asked with discipline. Our aim is
not to pin this ceiling to a single number, but to quantify how far it depends
on finite annotation, estimator choice, annotation noise, and the evaluation
protocol, and thereby to discipline how confidently saturation can be claimed.
We propose Bias-corrected Affective Ceiling Estimation (BACE), an analysis
framework that estimates a bias-corrected ceiling, separates irreducible from
reducible error, and disciplines the resulting claims. An anchored
Dirichlet-mixture empirical Bayes estimator, bracketed between plug-in and NSB,
recovers the human-consensus distribution; an annotator split, a noise
deconvolution, and a fixed claim gate then attribute error without circularity.
Methodologically, unconstrained point estimates place reachability anywhere from
$0.38$ to $1.03$, so saturation cannot be decided by any single estimator.
Substantively, the only assertion passing the claim gate is that at least about
$33$\% of a representative classifier's error on GoEmotions is irreducible, with
the same pattern recurring on offensiveness and irony.

\vspace{0.5em}
\end{abstract}

\section{Introduction}
\label{sec:intro}

Driven by advances in deep learning, emotion recognition from text has continued to improve on benchmarks, and the field has entered a stage in which powerful classifiers compete over marginal differences in performance. However, whether a ceiling exists on the accuracy that can be achieved, and how close current classifiers are to that ceiling, has rarely been asked.

Unless it is determined whether residual errors stem from limited model capacity or from ambiguity intrinsic to human judgment, one cannot tell whether a reported improvement is genuine progress or mere fine-tuning toward saturation. Emotion is subjective, and multiple annotators routinely assign different labels; this human label variation is a natural candidate for the irreducible error. While perspectivism treats such variation as signal \citep{plank2022,uma2021,davani2022dealing}, ChaosNLI \citep{nie2020chaosnli} and CIFAR-10H \citep{peterson2019human} use annotation entropy as an informal ceiling, and soft-label frameworks \citep{baan2022stop} address it, inter-annotator agreement is not an upper bound on model performance \citep{richie2022iaa}. Yet emotion annotation typically provides only about three annotators per instance, and the resulting instability of ceiling estimation under such sparse annotation has not been addressed.

Whether saturation can be claimed safely is obstructed by four challenges. First, finite-sample entropy estimation is downward-biased by concavity \citep{miller1955,paninski2003}, yet informal ceilings do not correct for it. Second, the ceiling shifts substantially with the estimator and can yield the opposite conclusions of ``room remaining'' and ``saturation'' for the same classifier, yet this variation has not been quantified. Third, building the ceiling and the evaluation from the same annotations introduces circularity, so the ceiling overfits to noise and can suggest superhuman performance. Fourth, observed disagreement conflates intrinsic ambiguity with annotation noise, and treating all of it as irreducible overestimates the ceiling and falsely suggests saturation.

In this study, Bias-corrected Affective Ceiling Estimation (BACE) is proposed. Its aim is not to estimate the ceiling as a single point, but to quantify how much ceiling estimation depends on finite annotation, estimator choice, noise, and the evaluation protocol, and thereby to discipline how far saturation can be claimed. BACE estimates the human consensus distribution and an information-theoretic ceiling (Estimate), separates classifier error into aleatoric and epistemic components without circularity (Separate), and admits only robust claims under a fixed discipline (Discipline). Through bias-corrected entropy estimation with systematic bracketing, annotator splitting, deconvolution, a claim gate, and predictability maps over emotion, offensiveness, and irony, it is shown that reachability depends heavily on the estimator, so that saturation cannot be asserted on its own, while a fixed fraction of the error nonetheless remains irreducible as a conservative lower bound. This shifts benchmark interpretation from a pursuit of absolute accuracy toward separating, in information-theoretic terms, the reducible room from the irreducible limit.

\section{Related Work}
\label{sec:related}

\subsection{Human Label Variation and Learning from Disagreement}
\label{sec:rel_disagreement}

Treating annotator disagreement in subjective tasks as signal rather than noise is now established. Plank \citep{plank2022} showed that human label variation is pervasive in NLP and questioned single-gold evaluation, and Uma et al. \citep{uma2021} surveyed learning from disagreement across hard-label aggregation, soft labels, and annotator modeling. Pavlick and Kwiatkowski \citep{pavlick2019} found disagreement in natural language inference to be intrinsic, and Aroyo and Welty \citep{aroyo2015truth} criticized the single-truth myth, noting that for subjective tasks such as emotion, aggregated hard labels discard essential information.

Perspectivist work further models annotator individuality: Davani et al. \citep{davani2022dealing} predicted individual labels before aggregation, the LeWiDi shared task \citep{leonardelli2023lewidi} standardized soft-label evaluation of disagreement, and reliability estimators such as MACE \citep{hovy2013} separated annotator noise from genuine ambiguity. These works preserve disagreement but neither quantify the ceiling it imposes on attainable accuracy nor address the finite-sample bias of estimating entropy from few annotations, which is where we depart by analyzing the estimator dependence of converting disagreement into a predictability ceiling.

\subsection{Bayes Error Ceilings and Entropy Estimation}
\label{sec:rel_ceiling}

The accuracy ceiling has long been studied in information theory. Fano's inequality \citep{fano1961,cover2006infotheory} links conditional entropy to a lower bound on any predictor's error, and for a known conditional distribution the tight $0$--$1$ lower bound is the complement of the maximum posterior probability. Recent direct Bayes-error estimation bypasses entropy: Ishida et al. \citep{ishida2023directbayes} estimated the binary Bayes error from class-uncertainty labels, and Ushio et al. \citep{ushio2025softbayes} extended this to soft labels and showed that calibration alone is insufficient. Richie et al. \citep{richie2022iaa} showed by simulation that inter-annotator agreement is not an upper bound on model performance, but offered no principled alternative.

Estimating entropy from finite samples is itself hard. The plug-in estimator is downward-biased by concavity \citep{miller1955}, as characterized by Paninski \citep{paninski2003}; the NSB estimator \citep{nemenman2002} is low-bias in undersampled regimes, with further options in coverage adjustment \citep{chao2003}, shrinkage \citep{hausser2009}, and a general Bayesian framework \citep{archer2014}, while Wolpert and Wolf \citep{wolpert1995} and Minka \citep{minka2000} provide closed-form and fixed-point Dirichlet--multinomial estimators. None of these has been applied systematically to the emotion-predictability ceiling, where informal ceilings often use uncorrected plug-in entropy. We instead integrate them into a systematic bracket via a hierarchical empirical Bayes over the cross-instance structure of emotion and, unlike direct Bayes-error estimation for the binary case, take an entropy and Fano path with explicit finite-sample correction for multi-class emotion; the direct estimator coincides with the plug-in lower end and so underestimates the bound (Section~\ref{sec:e1}).

\subsection{Aleatoric and Epistemic Uncertainty in Affective Computing}
\label{sec:rel_uncertainty}

Decomposing uncertainty into an irreducible aleatoric and a reducible epistemic component is widely adopted, with ChaosNLI \citep{nie2020chaosnli} in natural language inference and CIFAR-10H \citep{peterson2019human} in image classification both treating human disagreement as a lower bound on uncertainty. In affective computing, Baan et al. \citep{baan2022stop} evaluated model calibration against human uncertainty, while datasets provide multiply annotated resources: GoEmotions \citep{demszky2020} with annotator identifiers, SemEval emotion tasks \citep{mohammad2018semeval} and BRIGHTER \citep{muhammad2025brighter} for multilingual emotion, and MD-Agreement \citep{leonardelli2021}, EPIC \citep{frenda2023epic}, and MultiPICo \citep{casola2024multipico} for offensiveness and irony. Yet none organizes both systematic and sample uncertainty across emotion benchmarks while blocking circular reasoning and overclaiming, and some use only entropy estimation \citep{nemenman2002} or disagreement deconvolution \citep{gordon2021} in isolation. In contrast, we jointly provide instance-level entropy estimation, bias correction, rigorous Fano and Bayes ceilings, aleatoric/epistemic decomposition, and claim gating.

\section{Methodology}
\label{sec:method}

\subsection{Problem Definition}
\label{sec:problem_def}

Let $\{x_i\}_{i=1}^{M}$ denote the set of instances, $\mathcal{Y}$ the label space, and
$K=|\mathcal{Y}|$ its cardinality. Each instance $x_i$ is independently labeled by $n_i$
annotators, and the number of annotators who assign label $y$ is denoted by $c_{iy}$, where
$\sum_{y} c_{iy}=n_i$. The human-consensus distribution $p_i(y):=p(y\mid x_i)$ is
operationally defined as the probability that a single annotator drawn at random from
population $A$ assigns label $y$ under guideline $G$ and context $C$. Consequently, $p_i$ is
relative to the four-tuple $(A,G,C,\mathcal{Y})$, and this relativity is clarified in the
Limitations section.

The conditional entropy of instance $i$ and its dataset-level average are given respectively
by $H_i := -\sum_{y\in\mathcal{Y}} p_i(y)\log_2 p_i(y)$ and
$H(Y\mid X) := \frac{1}{M}\sum_{i=1}^{M} H_i$, where $H_i$ is the intrinsic ambiguity of
instance $x_i$.

Because a ceiling becomes meaningful only once the evaluation protocol is fixed, we define
three protocols separately. Under the random-annotator $0$--$1$ protocol P1, the ground truth
$Y$ is a single sample from $p_i$, and the error rate is
$P_e := \frac{1}{M}\sum_{i=1}^{M}\Pr(\hat{y}_i \neq Y_i)$. Under the majority-vote protocol P2,
the ground truth is the majority vote over the realized annotators, itself a random variable
under finite samples. Under the soft-label protocol P3, the model distribution $q_i$ is scored
by its cross-entropy to $p_i$. Because conflating protocols produces category inconsistencies,
every results table carries a protocol column.

When $p_i$ is known, an exact lower bound on the $0$--$1$ error under P1 is given directly
without invoking Fano's inequality, and
\begin{equation}
e_i^{*} := 1-\max_{y\in\mathcal{Y}} p_i(y),
\qquad
C := \frac{1}{M}\sum_{i=1}^{M} e_i^{*}
\label{eq:bayes}
\end{equation}
gives the predictability ceiling as the exact Bayes error, where $e_i^{*}$ denotes the
unreachable error lower bound at $x_i$ and $C$ its instance average. The classifier error
$\mathrm{Err}_{\mathrm{SOTA}}(P)$ is decomposed into the irreducible component $C(P)$ and the
reducible component $G(P):=\mathrm{Err}_{\mathrm{SOTA}}(P)-C(P)$, and the reachability is
\begin{equation}
R(P) := \frac{C(P)}{\mathrm{Err}_{\mathrm{SOTA}}(P)} \in (0,1] ,
\label{eq:reach}
\end{equation}
where $C(P)$ is aleatoric uncertainty, $G(P)$ is epistemic uncertainty, and $R(P)$ is the
irreducible fraction of the error. The central question is how stably $R(P)$ can be determined
from finite annotations.

\subsection{Overview of BACE}
\label{sec:overview}

\begin{figure*}[t]
  \centering
  \includegraphics[width=\textwidth]{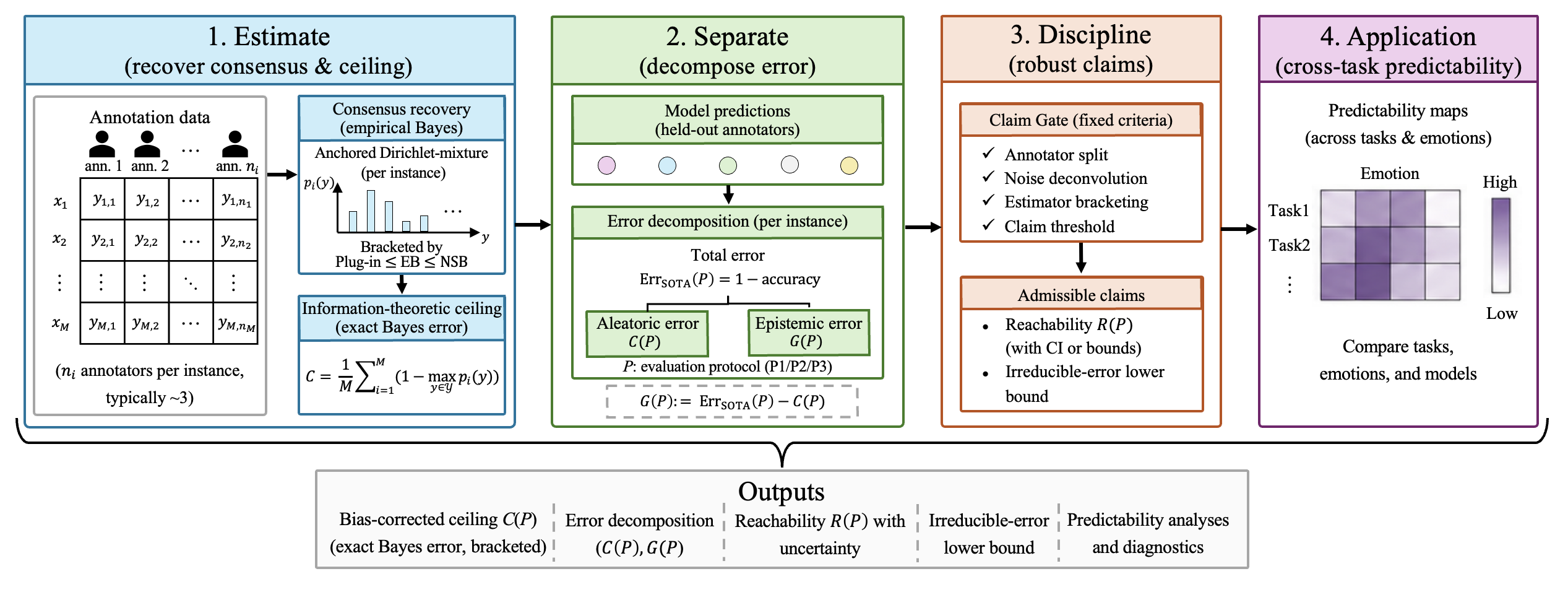}
  \caption{Overview of BACE: three core layers (Estimate, Separate, Discipline)
  followed by a cross-task application.}
  \label{fig:overview}
\end{figure*}

When $p_i$ is estimated from few annotations, the plug-in estimator biases the entropy
downward owing to concavity, reaching the order of $1$ bit for $n_i=3$ with an effective label
count of $5$ to $10$, so an uncorrected point estimate cannot serve as the primary estimator.
BACE is organized as three core layers, Estimate, Separate, and Discipline, followed by a
cross-task application (Figure~\ref{fig:overview}), and it handles this bias systematically
through four technical components detailed below: anchored Dirichlet-mixture empirical Bayes;
instance-level Fano and exact Bayes ceilings; aleatoric/epistemic decomposition based on
annotator splits; and deconvolution of annotation noise together with a claim gate. Together
these prevent circular reasoning and overclaiming while producing a cross-task predictability
map. By the relativity of the four-tuple in
Section~\ref{sec:problem_def}, these ceilings are predictability with respect to a specific
annotator population, observation channel, and taxonomy, not claims about the unknowability of
emotion itself.

\subsection{Anchored Dirichlet-Mixture Empirical Bayes}
\label{sec:mixeb}

The goal of this estimator is not entropy estimation per se, but the stable estimation of the downstream predictability ceiling and reachability. In what follows, the bias of entropy estimation is therefore controlled as a systematic error source that can distort the ceiling and reachability. The plug-in entropy carries a systematic bias $\mathbb{E}[\hat{H}^{\mathrm{plug}}]-H \approx -(K^{*}-1)/(2n\ln 2)$ \citep{miller1955,paninski2003}, where $K^{*}$ denotes the effective number of labels. Underestimating entropy inflates the epistemic gap, whereas overestimating it overclaims that the classifier has reached the ceiling, so these two dangerous directions point opposite to each other. Accordingly, every principal quantity is bracketed between the conservative plug-in lower side and the NSB \citep{nemenman2002} upper side rather than reported as a single point estimate.

The main estimator is an empirical Bayes that shares the inter-instance structure of the emotion label distribution as a hierarchical prior, transferring the base rate and confusion-pair structure learned from all instances to low-$n$ instances for which entropy is unrecoverable at $n_i=3$. A single asymmetric Dirichlet prior cannot express a mixture of near-unanimous and split instances and, on real data, exceeds the NSB upper end at low granularity, so a finite Dirichlet-mixture prior
\begin{equation}
\begin{aligned}
p_i &\sim \sum_{s=1}^{S} w_s\,\mathrm{Dir}(\tau_s m_s),\\
c_i &\sim \mathrm{Multinomial}(n_i, p_i)
\end{aligned}
\label{eq:mixprior}
\end{equation}
is adopted, where $w_s$ is the component weight, $m_s\in\Delta^{K-1}$ is the base measure of component $s$, and $\tau_s>0$ is its concentration. Assignment uses the responsibility $r_{is}\propto w_s\,\mathrm{DM}(c_i\mid \tau_s m_s)$, where $\mathrm{DM}$ denotes the Dirichlet--multinomial marginal likelihood, so leakage from double use of the observed counts does not arise structurally. Closed forms for the posterior entropy and $\max_y p_{iy}$, the Minka-type fitting of $(\tau_s,m_s)$ \citep{wolpert1995,minka2000}, category-wise anchoring, and the holdout selection of $S$ are given in Appendix~\ref{app:mixeb}. The ordering $\mathrm{plug\text{-}in}\le\mathrm{EB}\le\mathrm{NSB}$ is a diagnostic expectation rather than a theorem; when a reversal persists even after correction, the empirical Bayes point is demoted to a reference value and claims are made only at both ends of the systematic bracket.

\subsection{Fano and Exact Bayes Ceilings}
\label{sec:fano}

For a deterministic predictor $\hat{y}=g(x)$ forming the Markov chain $Y$--$X$--$\hat{Y}$, Fano's inequality \citep{fano1961,cover2006infotheory} gives, per instance,
\begin{equation}
H_i \le h_b(e_i) + e_i\log_2(K-1),
\qquad
e_i \ge f_K(H_i) ,
\label{eq:fano}
\end{equation}
where $h_b$ is the binary entropy, $e_i:=\Pr(\hat{y}_i\neq Y_i\mid x_i)$ is the instance error, and $f_K:=\varphi_K^{-1}$ is the convex increasing inverse of $\varphi_K(t):=h_b(t)+t\log_2(K-1)$. The Fano lower bound is always looser than the exact Bayes error, so the main ceiling for P1 is taken to be the exact Bayes error of Eq.~\eqref{eq:bayes}, and the Fano-type lower bound is reported alongside it in three complementary roles. These include unifying $0$--$1$ error and soft-label evaluation through $\mathbb{E}[\mathrm{CE}]\ge H(Y\mid X)$ (Appendix~\ref{app:fano}).

For multi-label emotion, the joint distribution has $2^{K}$ configurations and cannot be estimated from $n_i\le 5$, so only the marginal quantities required by the evaluation metrics are estimated. Under the binary decomposition ($Y_k=1[k\in S]$, $p_{ik}:=\Pr(Y_k=1\mid x_i)$), $\log_2(K-1)=0$, so Fano degenerates to the entropy condition $h_b(e)\ge H$, the lower inverse becomes exactly $e_{ik}\ge\min(p_{ik},1-p_{ik})$, and the entropy-based lower bound coincides with the exact Bayes error, so the per-emotion ceiling is exact. Its instance-wise lower bound is obtained in closed form through a per-label Beta--Binomial empirical Bayes prior and the regularized incomplete beta function (Appendix~\ref{app:fano}). F1 is not decomposable and its Bayes-optimal value depends on the joint distribution, so what is reported is the oracle plug-in F1 obtained by optimal thresholding of the marginal probabilities \citep{ye2012}, which is a marginal reference value rather than a proof of the upper bound.

\subsection{Aleatoric/Epistemic Decomposition via Annotator Splitting}
\label{sec:decomp}

Constructing both the ceiling and the evaluation from the same annotations introduces the circular reasoning that the ceiling overfits to annotation noise. To prevent this, the annotators of each instance are split into two disjoint groups. That is, the analytic ceiling and the oracle predictor are constructed from $\hat{p}_i^{A}$ on the estimation split A, and the classifier under evaluation and the oracle are scored on the evaluation split B, where $|A_i|=\lceil n_i/2\rceil$, $|B_i|=\lfloor n_i/2\rfloor$, and an odd extra annotator is assigned to A in order to prioritize the estimation bottleneck. This split is made deterministic by a fixed-seed permutation over the lexicographically sorted record table (Appendix~\ref{app:boot}).

The reachability $R(P)$ of Eq.~\eqref{eq:reach} is taken as the principal reported quantity, and its irreducible component $C(P)$ is a lower bound that applies to any model predicting labels of the same input channel and annotator population. For protocol consistency, the ceiling is computed under the same protocol as the classifier. As a diagnostic, the oracle constructed from A and scored on B agrees with the analytic ceiling, supporting the estimation. When the classifier falls below the ceiling beyond the confidence interval ($\hat{G}<0$), this is treated, by a fixed rule, as an alarm for underestimation of the ceiling, leakage of the test annotations, or protocol mismatch, mechanically ruling out the erroneous conclusion of superhuman performance.

\subsection{Annotation-Noise Deconvolution and the Claim Gate}
\label{sec:deconv}

Because the observed disagreement mixes true ambiguity and annotation error, the entropy of the raw $\hat{p}$ overestimates ambiguity. Following the perspectivist standpoint \citep{plank2022,uma2021}, we do not remove it completely but fit a lightweight model in which annotator $a$ reports the signal with probability $1-\varepsilon_a$ and a granularity-matched noise $\nu$ with probability $\varepsilon_a$,
\begin{equation}
\Pr(y_t=y) = (1-\varepsilon_{a_t})\,\tilde{p}_{i_t}(y) + \varepsilon_{a_t}\,\nu(y) ,
\label{eq:deconv}
\end{equation}
by EM \citep{gordon2021}. Here $\nu$ is restricted to uniform and base-rate types, with no confusion matrix that would absorb true ambiguity, and $\varepsilon_a$ (shared across annotators) and $\tilde{p}_i$ (shared across instances) are identified from the crossed structure. Monotonicity is not assumed: denoising lowers the ceiling for plug-in but, for mixture empirical Bayes and NSB, can raise it above the raw data through the interaction with finite-sample correction or hierarchical smoothing, so both conditions are reported and the ordering depends on the estimator. Only the post-deconvolution plug-in serves as the conservative lower bound, where denoising and the plug-in downward bias compound to the smallest reachability, while the post-deconvolution mixture empirical Bayes and NSB are a sensitivity analysis. The EM updates, the reuse of $\hat{\varepsilon}_a$ across granularities, and the recovery check are in Appendix~\ref{app:deconv}.

The claim gate disciplines claims a priori. A reachability claim that the classifier attains $X$\% of the ceiling is asserted only if it holds at the conservative ends of both the systematic bracket and the $95$\% confidence interval, at both granularities, and in particular at the post-deconvolution plug-in end, where the ceiling is smallest and reachability hardest to attain; otherwise it is reported as an interval without a point claim. The gate thereby secures, at the methodological level, the central claim that an informal human ceiling moves substantially under the choice of estimator.

\section{Experiments}
\label{sec:experiments}

\subsection{Datasets and Experiment Design}
\label{sec:design}

Our primary foundation is GoEmotions \citep{demszky2020}, which fully releases annotator-identified raw annotations over $27$ emotions plus neutral, letting us relate the granularity of the emotion space to the ceiling. For robustness we also use the binary offensiveness task MD-Agreement \citep{leonardelli2021}, the binary irony task EPIC \citep{frenda2023epic}, and the English portion of BRIGHTER \citep{muhammad2025brighter} (statistics in Appendix~\ref{app:datasets}). GoEmotions has only $3.41$ annotators per instance and base rates spanning two orders of magnitude, from $0.262$ for neutral to $0.003$ for grief, so the per-instance entropy is noisy and a hierarchical empirical Bayes bias correction is indispensable.

The evaluated classifier is the representative public SamLowe/roberta-base-go\_emotions, which we do not claim to be the single best model; bhadresh-savani/bert-base-go-emotion and monologg/bert-base-cased-goemotions-original are also evaluated in Section~\ref{sec:e2} against the same ceiling and instances. The ceiling estimators plug-in, Miller--Madow \citep{miller1955}, NSB \citep{nemenman2002}, and the proposed mixture empirical Bayes share a common implementation, and the direct and soft-label Bayes-error estimators \citep{ishida2023directbayes,ushio2025softbayes} and bias-corrected entropy estimators \citep{chao2003,hausser2009} serve as external baselines on synthetic distributions with known entropy (Appendix~\ref{app:estcomp}, \ref{app:tests}). We report the exact Bayes error for P1, lower bounds for Hamming and macro/micro-F1, the cross-entropy lower bound for P3, and the reachability $R$, with bootstrap intervals separated from the systematic estimator bracket. To prevent leakage, the decomposition uses only the official GoEmotions test instances and the prior is fitted on the out-of-test A-side annotations.

\subsection{The Need for a Bias-Corrected Bracket}
\label{sec:e1_bracket}

Table~\ref{tab:goemo} shows the GoEmotions coarse-granularity ceilings under plug-in, mixture empirical Bayes, and NSB, together with the deconvolved plug-in end and the A/B-track decomposition. At the L2 Ekman $7$-class level, the systematic bracket spans $[0.258, 0.391]$, confirming on real data that a single point estimate swings the result by roughly $13$ points in error rate. Mixture empirical Bayes recovers the ordering $\mathrm{plug\text{-}in}\le\mathrm{EB}\le\mathrm{NSB}$ at L2, but a reversal remains at L1, so the empirical Bayes point is demoted to a reference value (Section~\ref{sec:mixeb}). The deconvolved plug-in end is $0.197$ at L2 and $0.179$ at L1, providing the conservative end for reachability claims. As the claim gate requires, the all-annotation track and the A/B tracks agree within the systematic range. The differences among these estimators are not mere numerical fluctuation: since the model error is fixed, where the ceiling is placed directly determines whether the same classifier appears far from the ceiling or already at it.

\begin{table*}[t]
  \centering
  \small
  \setlength{\tabcolsep}{8pt}
  \caption{Predictability ceilings and aleatoric/epistemic decomposition for GoEmotions (P1 exact Bayes error).}
  \label{tab:goemo}
  \begin{tabular}{lccccccc}
    \toprule
    Gran. & plug-in & NSB & dec.$\times$plug & $\mathrm{Err}_{\mathrm{SOTA}}$ & ceiling & gap & $R$ cons.\ [CI] \\
    \midrule
    L2 ($K{=}7$) & 0.258 & \textbf{0.391} & 0.197 & 0.388 & 0.150 & 0.238 & \textbf{0.386} [.368,.403] \\
    L1 ($K{=}4$) & 0.227 & \textbf{0.328} & 0.179 & 0.350 & 0.134 & 0.216 & \textbf{0.383} [.366,.402] \\
    \bottomrule
  \end{tabular}
\end{table*}

\subsection{Exact Lower Bounds from Binary Decomposition and Per-Label Predictability}
\label{sec:e1}

Under the binary decomposition, the sample deficiency vanishes for $n\ge 2$, the lower bound becomes an exact Bayes error with zero slack, and the Hamming, macro, and micro lower bounds coincide at $0.0369$. The existing direct and soft-label estimators \citep{ishida2023directbayes,ushio2025softbayes} reduce to the lower plug-in end and underestimate this bound (Appendix~\ref{app:estcomp}). The per-label irreducible-error lower bound $e^{*}_k$ ranges from $0.0033$ for grief to $0.218$ for the frequent and ambiguous neutral, giving an independent map of which emotions are in principle hard to predict. The bounds for all $28$ labels and the oracle plug-in F1 are given in Appendix~\ref{app:perlabel}.

\subsection{Estimator Dependence of Reachability and a Robust Lower Bound}
\label{sec:e2_reach}

Having established that the ceiling itself depends on the estimator, we verify whether this dependence affects the substantive conclusion of saturation. Table~\ref{tab:goemo} decomposes the error of the evaluated classifier into irreducible and reducible components. At the default smoothing $\beta=0.7$, the conservative reachability is about $0.38$ at both granularities (L2 $0.386$, L1 $0.383$), and its minimum across $\beta\in\{0.5,0.7,0.9\}$ and split seeds falls to about $0.34$ ($0.3375$ at L1 with $\beta=0.5$; Appendix~\ref{app:boot}). The only statement that passes the claim gate is therefore that at least about $33$\% of the P1 error on GoEmotions is an irreducible aleatoric component, and this is a conservative lower bound rather than a point estimate. Because the number of observed annotators is small, the denoising effect and the finite-sample bias correction cannot be fully separated, so this lower bound is interpreted on the conservative side (Appendix~\ref{app:deconv}).

\begin{figure*}[t]
  \centering
  \includegraphics[width=0.85\textwidth]{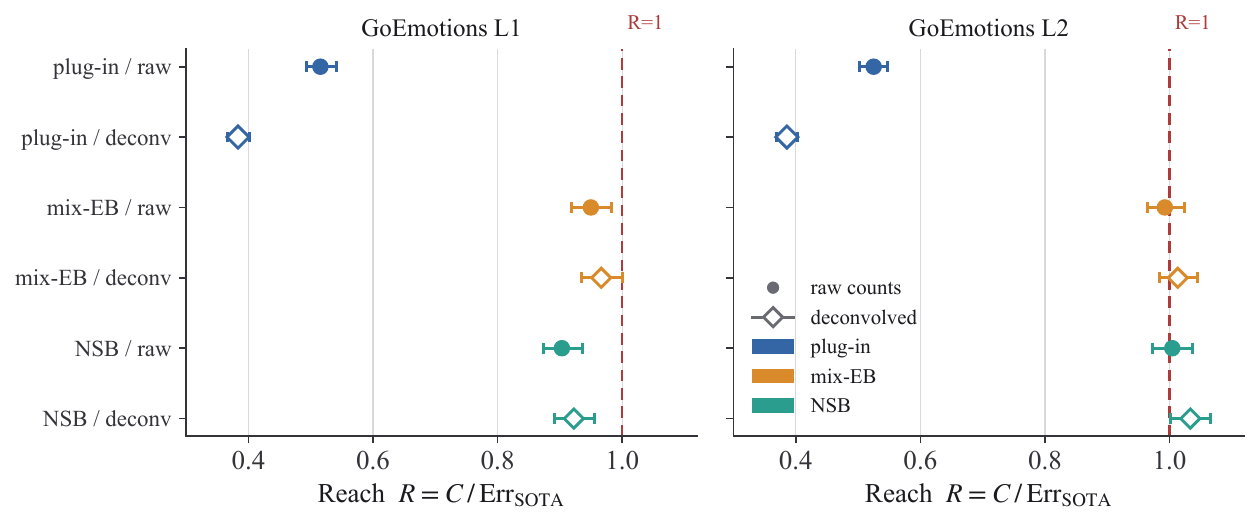}
  \caption{The reachability $R$ depends on the combination of estimator, noise handling, and finite-sample correction, and the direction of its change is not constant.}
  \label{fig:reach}
\end{figure*}

Figure~\ref{fig:reach} visualizes how the reachability $R$ depends on the choice of estimator and deconvolution. For the same evaluation error, the reachability swings from $0.38$ at the conservative deconvolved plug-in end, through about $0.52$ at the raw plug-in end, to between $0.90$ and $1.03$ at the mixture empirical Bayes and NSB ends. The effect of deconvolution is not one-directional: it lowers the reachability for plug-in but raises it for mixture empirical Bayes and NSB, so the direction of change differs by estimator. It is thus the core of this work that the verdict of saturation holds only at the empirical Bayes and NSB ends and fails at the plug-in end. The binary F1 and the soft-label P3 likewise span a wide range (Appendix~\ref{app:perlabel}). The diagnostic quantity $\hat{G}$ is positive at the conservative ends and turns only slightly negative at the L2 upper-end NSB (raw counts) and the deconvolved mixture empirical Bayes and NSB, which is an alarm rather than superhuman performance.

\subsection{Dependence on the Evaluated Model (Multi-Model)}
\label{sec:e2}

Table~\ref{tab:multimodel} reports the reachability of four fine-tuned classifiers (one of which is ModernBERT-large) and one GPT-4o-mini few-shot classifier, decomposed against the same L2 ceiling, and shows that the verdict of saturation also changes with the classifier placed in the denominator. The raw empirical Bayes reachability spans $[0.732, 1.009]$ and crosses $1$ through the denominator alone, so a larger backbone does not break the ceiling and no single value of $R$ decides the question of saturation. The details of the multi-model decomposition are given in Appendix~\ref{app:multimodel}.

\begin{table*}[t]
  \centering
  \small
  \setlength{\tabcolsep}{8pt}
  \caption{Multi-model decomposition against the same GoEmotions L2 ceiling (P1).}
  \label{tab:multimodel}
  \begin{tabular}{lccc}
    \toprule
    Classifier & $\mathrm{Err}$ & $R_{\mathrm{cons}}$ [CI] & $R_{\mathrm{EB}}$ [CI] \\
    \midrule
    bhadresh-BERT    & 0.382 & 0.392 [.375,.411] & 1.009 [.977,1.043] \\
    ModernBERT-large & 0.386 & 0.388 [.371,.405] & 0.997 [.967,1.028] \\
    SamLowe-RoBERTa  & 0.388 & 0.386 [.368,.403] & 0.992 [.964,1.023] \\
    monologg-BERT    & 0.419 & 0.357 [.342,.372] & 0.918 [.891,.946] \\
    GPT-4o-mini (fs) & 0.526 & 0.285 [.273,.297] & 0.732 [.714,.750] \\
    \bottomrule
  \end{tabular}
\end{table*}

\subsection{Cross-Task Predictability Map}
\label{sec:map}

\begin{figure}[t]
  \centering
  \includegraphics[width=\columnwidth]{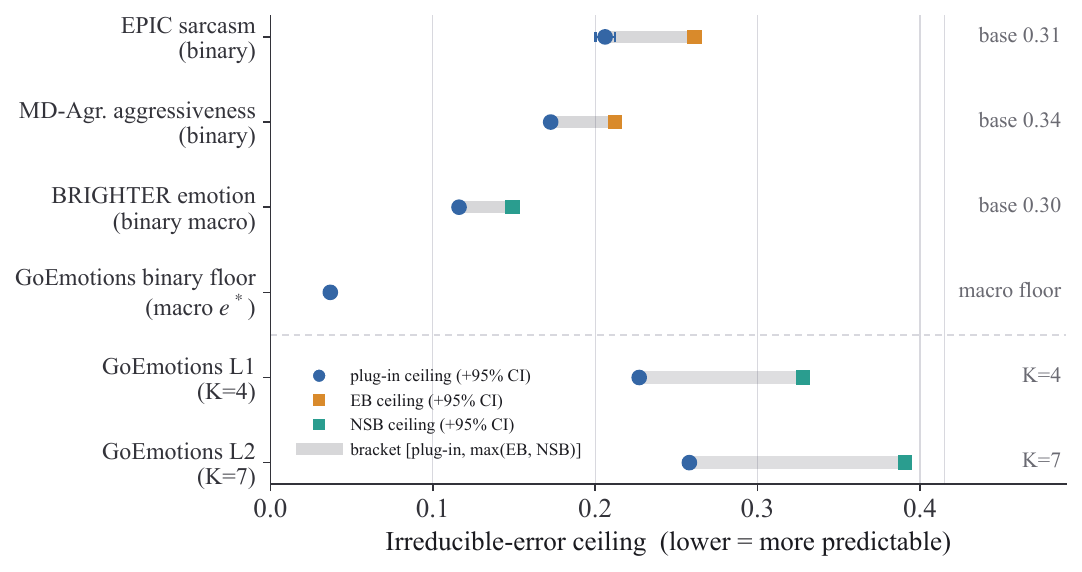}
  \caption{Cross-task predictability map controlling for base rate.}
  \label{fig:map}
\end{figure}

Figure~\ref{fig:map} shows the cross-task predictability map across emotion, offensiveness, and irony, controlling for base rate as cross-task comparison requires. At a base rate of about $0.3$, the irreducible ambiguity is largest for irony at $[0.206, 0.261]$, exceeding that of offensiveness and emotion presence. The lower bound also rises monotonically with the number of classes. The deconvolved conservative ends are $0.139$ for irony and $0.137$ for offensiveness, and the mean noise rate is largest for irony at $0.257$ and $0.151$ for offensiveness. The details of deconvolution and BRIGHTER are given in Appendix~\ref{app:perlabel}.

Table~\ref{tab:crosstask} decomposes an off-the-shelf public classifier on each corpus against its conservative ceiling and shows that two core findings recur across tasks. Namely, the conservative irreducible component remains positive for offensiveness, irony, and emotion, and the reachability again depends strongly on the estimator. Hence the positive $C$ and this estimator dependence are task-general rather than an artifact of GoEmotions. However, because these classifiers are not fine-tuned on the target corpora, these reachabilities are lower bounds. The details are given in Appendix~\ref{app:crosstask}.

\begin{table*}[t]
  \centering
  \small
  \setlength{\tabcolsep}{8pt}
  \caption{Cross-task SOTA decomposition against the conservative ceiling.}
  \label{tab:crosstask}
  \begin{tabular}{lccccc}
    \toprule
    Task & $\mathrm{Err}_{\mathrm{SOTA}}$ & $C_{\mathrm{cons}}$ & $R_{\mathrm{cons}}$ [CI] & $R_{\mathrm{EB}}$ & $R_{\mathrm{NSB}}$ \\
    \midrule
    Offensiveness (MD-Ag.)  & 0.332 & 0.137 & 0.413 [.404,.423] & 0.640 & 0.618 \\
    Irony (EPIC)            & 0.388 & 0.139 & 0.359 [.342,.375] & 0.672 & 0.603 \\
    Emotion (BRIGHTER)      & 0.275 & 0.116 & 0.422 [.416,.429] & 0.482 & 0.543 \\
    \bottomrule
  \end{tabular}
\end{table*}

\section{Discussion}
\label{sec:discussion}

\subsection{Reachability Is Estimator-Dependent}
\label{sec:disc_reach}

The central finding is that the informal ceiling depends very strongly on the
estimator, so a plateau judgment cannot be entrusted to a single point estimate.
For the same evaluation error, reachability swings from $0.38$ to $1.03$, and
even with the ceiling and estimator fixed it flips across $1$ depending on which
strong public classifier is the denominator (Section~\ref{sec:e2}). Hence the
only assertion passing the claim gate is that at least about $33$\% of the
GoEmotions evaluation error is irreducible, a lower bound the systematic bracket
states explicitly as an interval, and this estimator dependence recurs on
offensiveness, irony, and emotion (Appendix~\ref{app:crosstask}).

\subsection{What Is Irreducible in Emotion}
\label{sec:disc_irreducible}

The per-label lower bounds differ greatly, ranging from below $0.01$ for rare,
high-agreement emotions to $0.218$ for neutral (Section~\ref{sec:e1}). This is
internal structure that average accuracy conceals. Once base rates are
controlled for, irony carries the largest irreducible ambiguity; however, the
lower bounds for irony and offensiveness nearly converge after deconvolution, so
much of the raw disagreement is annotation noise, and treating disagreement
directly as irreducible ambiguity overestimates the ceiling. Predictability is
thus structured by label, task, and granularity, and because the lower bound
rises with the number of classes, the design of the taxonomy itself governs it.

\subsection{Implications for Benchmarking and Annotation Budgets}
\label{sec:disc_implications}

Because reachability depends on the estimator, a leaderboard's residual error is
not uniformly a reducible epistemic gap, and since at least about $33$\% of the
GoEmotions error is irreducible, the room for improvement is often overestimated.
Benchmark reports should therefore position achieved accuracy relative to an
irreducible lower bound with a systematic bracket. Since the bracket width
depends strongly on the number of annotators, about $0.05$ for BRIGHTER and
$0.10$ for the sparse L1 granularity of GoEmotions, designers can set annotators
per instance from a target precision, and released identifiers let deconvolution
separate genuine ambiguity from annotation noise.

\section{Conclusion}
\label{sec:conclusion}

We addressed with BACE the accuracy race in emotion recognition that proceeds while the
predictability ceiling remains unknown. Bracketing an anchored Dirichlet-mixture empirical
Bayes between the plug-in and NSB, and decomposing error through an annotator split, a
deconvolution, and a fixed claim gate, BACE estimates a bias-corrected ceiling from human
label variation and blocks circular reasoning and over-claiming.

Reachability swings from $0.38$ to $1.03$ with the estimator and the evaluated classifier, so
a plateau verdict is impossible without estimator discipline. The only statement passing the
claim gate is that at least about $33$\% of the evaluation error on GoEmotions is irreducible,
a pattern that recurs on offensiveness and irony. In future work, we will extend the framework
to more languages and taxonomies.

\section*{Limitations}

Our ceilings are relative to a specific annotator population, observation channel, and taxonomy: a model with extra-textual information is not a counterexample to Fano, and the ceiling concerns this pool's label distribution, not the unknowability of emotion. In the L1 sentiment setting the estimator ordering reversal persists after correction, so the mixture empirical Bayes point is used only as a reference and claims are made at the two ends of the systematic bracket. At the observed annotator counts, deconvolution denoising cannot be separated from finite-sample bias, so the deconvolved plug-in end serves only as a conservative lower bound and the reported $33$\% is specific to GoEmotions; relatedly, $\hat{G}<0$ occurs only at the upper-end L2 ceilings and signals overestimation or leakage, not superhuman performance. Annotator-level overlap may remain because classifiers can see the same annotators' labels in training, and BRIGHTER carries no annotator identifiers, so only its raw-data ceiling is reported. Finally, a single LLM (GPT-4o-mini, one few-shot configuration) was evaluated, so the capability band is not exhaustive. All numbers are reported as they are.

\section*{Ethical Considerations}

Because this study addresses emotion labels that are subjective and can be
contested, several ethical considerations are made explicit. First, our
framework treats disagreement among annotators as signal rather than noise, and
avoids collapsing perspectival differences into a single majority-vote label.
Second, the estimated ceiling is a quantity relative to a specific annotator
population, annotation guideline, observation channel, and taxonomy, and is not
a claim about any individual's ``true emotion.'' Using our ceiling to determine
the emotional state of a particular person is therefore neither intended nor
appropriate.

Regarding data, only publicly released datasets were used, in accordance with
their respective licenses. GoEmotions is released under Apache-2.0, MD-Agreement
and EPIC under non-commercial licenses, and BRIGHTER is restricted to research
use. Our code is available at
\url{https://github.com/keito-git/affectceiling-bace}; the datasets
are not bundled with it, and retrieval scripts are distributed instead. No new human-subject data collection was
conducted, no personally identifying information was added, and no annotator was
re-identified.

A foreseeable misuse is to interpret a predictability ceiling as a fundamental
limit on the ability to read an individual's emotions. To guard against this,
the ceiling is presented at the level of datasets and populations and as a
conservative lower bound. This conservative framing is intended to curb
over-claiming in affective computing and to encourage an evaluation culture that
honestly reports where achieved accuracy stands relative to an irreducible
floor.

\bibliography{references}

\appendix
\section{Dataset Statistics}
\label{app:datasets}

Table~\ref{tab:datasets} summarizes the statistics of the datasets used in Section~\ref{sec:design}.

\begin{table*}[t]
  \centering
  \small
  \setlength{\tabcolsep}{6pt}
  \caption{Dataset statistics.}
  \label{tab:datasets}
  \begin{tabular}{lrrll}
    \toprule
    Dataset & Inst. & Ann. & $n_i$ (mean/med.) & Labels \\
    \midrule
    GoEmotions   & 57{,}828 & 82  & 3.41 / 3 & multi-label 27+neutral \\
    MD-Agreement & 10{,}753 & 819 & 5 / 5    & binary offensiveness \\
    EPIC         & 3{,}000  & 74  & 4.72 / 5 & binary irony \\
    BRIGHTER eng & 5{,}655  & no ID & 8.52 / 8 & 6 emotions, ordinal \\
    \bottomrule
  \end{tabular}
\end{table*}

\section{Comparison with Existing Bayes-Error Estimators}
\label{app:estcomp}

Table~\ref{tab:estcomp} relates the proposed bracket to the existing Bayes-error and entropy estimators of Section~\ref{sec:e1}. For the binary GoEmotions floor, which is the macro floor and equals the Hamming and cell micro floors, the direct estimator of Ishida et al.\ \citep{ishida2023directbayes} and the simplified soft-label isotonic estimator of Ushio et al.\ \citep{ushio2025softbayes}, whose calibration is nearly the identity map on empirical soft labels, both agree with the plug-in lower end of $0.0274$. In contrast, the bias-corrected Beta--Binomial point rises to $0.0369$ and the NSB upper end rises to $0.112$, so the direct estimator inherits the finite-sample downward bias and underestimates the floor. On the categorical side, the mixture empirical Bayes and NSB ceilings lie inside the range spanned by the coverage-adjusted estimator of Chao and Shen \citep{chao2003} and the shrinkage estimator of H\"{a}usser and Strimmer \citep{hausser2009}, where these informal ceilings are read off through the Fano inverse and the shrinkage estimator lies closest to NSB.

\begin{table*}[t]
  \centering
  \small
  \setlength{\tabcolsep}{8pt}
  \caption{Comparison with existing Bayes-error and entropy estimators.}
  \label{tab:estcomp}
  \begin{tabular}{llc}
    \toprule
    Block & Estimator & Value \\
    \midrule
    Binary floor & Ishida direct \citep{ishida2023directbayes} & 0.0274 \\
    Binary floor & Ushio soft-label \citep{ushio2025softbayes} & 0.0274 \\
    Binary floor & BACE Beta-Binomial EB (point) & \textbf{0.0369} \\
    Binary floor & BACE NSB $K{=}2$ (upper) & 0.1120 \\
    \midrule
    L2 $H$ / ceil. & plug-in (informal) & 0.713 / 0.259 \\
    L2 $H$ / ceil. & Chao--Shen \citep{chao2003} & 1.062 / 0.205 \\
    L2 $H$ / ceil. & H\"{a}usser shrink.\ \citep{hausser2009} & 1.547 / 0.385 \\
    L2 $H$ / ceil. & BACE mixture EB & 1.327 / 0.379 \\
    L2 $H$ / ceil. & BACE NSB (upper) & 1.393 / 0.390 \\
    \bottomrule
  \end{tabular}
\end{table*}

\section{Multi-Model Reachability}
\label{app:multimodel}

\begin{figure*}[t]
  \centering
  \includegraphics[width=\textwidth]{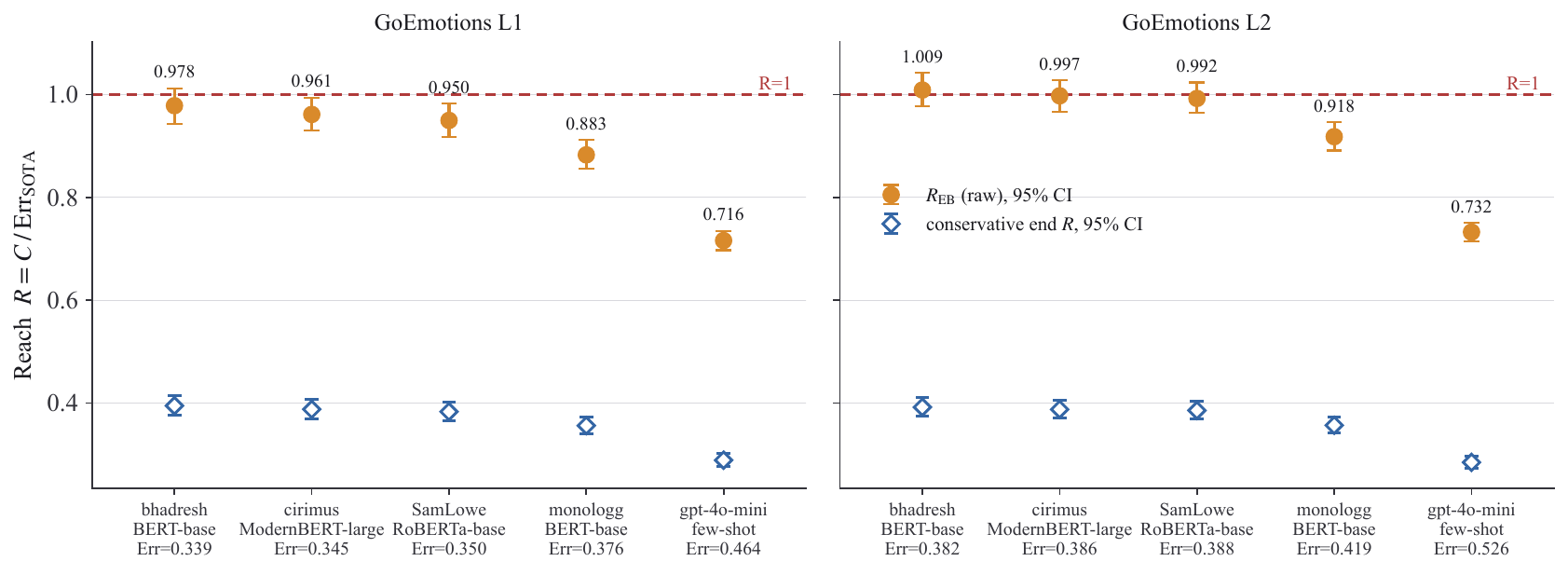}
  \caption{Multi-model reachability at L2.}
  \label{fig:multimodel}
\end{figure*}

The multi-model decomposition of Section~\ref{sec:e2} (Table~\ref{tab:multimodel}) decomposes four fine-tuned public classifiers spanning three base-size transformers and one ModernBERT-large, together with a GPT-4o-mini few-shot classifier belonging to a different capability class, against the same GoEmotions ceiling and the same $5{,}330$ evaluation instances, and Figure~\ref{fig:multimodel} visualizes their reachability. At the conservative end all five classifiers fall below $1$, yet at the raw empirical Bayes end the saturation verdict flips on the classifier alone. Specifically, the interval of bhadresh-BERT straddles $1$ whereas the interval of monologg-BERT lies entirely below $1$, and this holds even with the ceiling and the estimator held fixed. The larger ModernBERT-large backbone does not escape this cluster either. It attains essentially the same error as the base sizes ($0.386$), a raw empirical Bayes reachability of $0.997$ and an NSB reachability of $1.009$, placing it inside the same $R\approx 1$ saturation band as the base-size transformers. Enlarging the backbone therefore does not lower the error on the collapsed categories, and the base-size fine-tuned classifiers already sit at the reachability ceiling. Whether a stronger model breaks the ceiling is thus answered negatively within this range. The GPT-4o-mini classifier (openai/gpt-4o-mini, via OpenRouter, temperature $0$, an $8$-shot few-shot prompt with JSON output, no parsing failures across all $5{,}330$ instances) is the weakest on the $28$-class task and hence has the largest error, so under the same ceiling it attains the smallest reachability, pushing the raw empirical Bayes band down to $0.732$ and the conservative band down to $0.285$. Reachability therefore spans a broad capability range from fine-tuned transformers to LLM few-shot, and no single value of $R$ decides saturation.

\section{Mixture Empirical Bayes: Closed Form and Fitting}
\label{app:mixeb}

The instance-level posterior quantities of the anchored Dirichlet mixture empirical Bayes of Section~\ref{sec:mixeb} are obtained as a responsibility-weighted sum of per-component closed forms \citep{wolpert1995}. In particular, the posterior entropy is given by
\begin{equation}
\mathbb{E}[H_i\mid c_i] = \sum_{s=1}^{S} r_{is}\,\mathbb{E}_{\mathrm{Dir}(c_i+\tau_s m_s)}[H] ,
\label{eq:eb_h}
\end{equation}
where each term on the right-hand side denotes the expected entropy under the posterior Dirichlet of component $s$, evaluated by the Wolpert--Wolf closed form. The expectation of $\max_y p_{iy}$ entering the exact Bayes error is evaluated by allocating Monte Carlo samples in proportion to the responsibilities $r_{is}\propto w_s\,\mathrm{DM}(c_i\mid \tau_s m_s)$.

\begin{figure*}[t]
  \centering
  \includegraphics[width=0.6\textwidth]{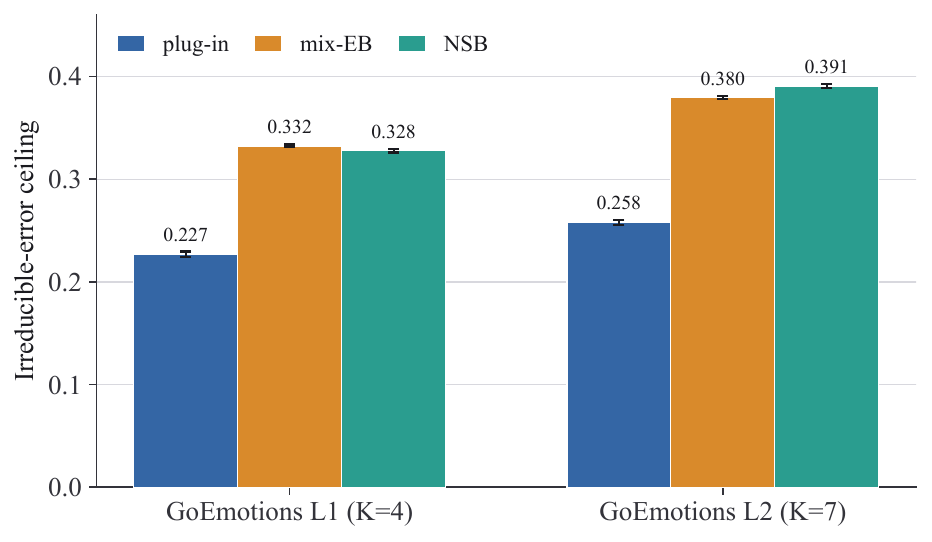}
  \caption{Ceiling on the irreducible error of GoEmotions under the plug-in, mixture empirical Bayes, and NSB estimators.}
  \label{fig:ceiling}
\end{figure*}

The hyperparameters $(w_s,\tau_s,m_s)$ are fitted by marginal-likelihood maximization, and the concentration and base measure of each component are obtained by Minka-style fixed-point iteration \citep{minka2000}. To avoid divergence on nearly binary labels for which the shared Minka update drives $\tau_s\to\infty$, a locally robust fit is used with an upper bound $\tau_s\le 10^{4}$ and a non-finite guard. Model selection over the number of components starts from $K$ components anchored to each category and one global splitting component, and $S$ is adopted only when it improves the mean holdout log marginal likelihood over $S=1$. On a synthetic population with peaked and diffuse modes, this holdout criterion improves the log marginal likelihood by $+0.111$ over $S=1$, and on GoEmotions the selected value is $S=8$ at L2 and $S=5$ at L1, both equal to $K+1$. Figure~\ref{fig:ceiling} visualizes the ceiling on the irreducible error of GoEmotions under the three estimators and shows that it is the estimator, not the data, that sets the target.

\section{Fano Bound: Three Roles and the Per-Instance Limit}
\label{app:fano}

At the dataset level, the per-instance Fano inequality of Eq.~\eqref{eq:fano} gives $P_e \ge (1/M)\sum_i f_K(H_i) \ge f_K(H(Y\mid X))$, where the second inequality follows from Jensen's inequality applied to the convex inverse $f_K$, so that applying the lower bound per instance before averaging is tighter than a single application to the aggregated entropy.

The exact Bayes error of Eq.~\eqref{eq:bayes} is the primary ceiling for P1, but the looser Fano-type lower bound is reported alongside it in three complementary roles. First, it makes explicit a model-agnostic guarantee, namely a theoretical skeleton that lower-bounds any future model. Second, against the upward plug-in bias of $\max_y p$, it provides an independent estimation path $f_K(\hat{H}_i)$ via entropy, used to diagnose the discrepancy between the two paths. Third, through the exact lower bound on log loss $\mathbb{E}[\mathrm{CE}]\ge H(Y\mid X)$, it unifies with P3, since the same conditional entropy ties $0$--$1$ error and soft-label evaluation into a single quantity.

In the binary decomposition of Section~\ref{sec:fano}, Fano degenerates to $e_{ik}\ge\min(p_{ik},1-p_{ik})$ and the entropy-based lower bound equals the exact Bayes error. The instance-level floor for binary label $k$ is then given in closed form by
\begin{multline}
\mathbb{E}[\min(p,1-p)\mid c] = \\
\mu\,I_{1/2}(a'+1,b') + (1-\mu)\bigl(1-I_{1/2}(a',b'+1)\bigr) ,
\label{eq:binclosed}
\end{multline}
where $I$ is the regularized incomplete beta function, $a'=c_{ik}+a_k$, $b'=n_i-c_{ik}+b_k$, and $\mu=a'/(a'+b')$, and where $(a_k,b_k)$ denotes the per-label Beta--Binomial empirical Bayes prior.

\section{Annotator-Split and Bootstrap Details}
\label{app:boot}

The A/B split of Section~\ref{sec:decomp} is made deterministic as follows. The annotation records are sorted lexicographically by instance identifier and annotator identifier, the annotators of each instance are permuted with a single fixed-seed random number generator, and the resulting split table is stored so that all subsequent experiments refer only to this table. The frozen split uses seed $20260723$. Its structural A ratio is $0.613$, which arises from assigning the odd extra annotator to A, and the per-annotator A ratio has mean $0.593$ and standard deviation $0.078$.

Confidence intervals are obtained by instance-resampling bootstrap with the prior fit and deconvolution held fixed, because fully refitting the mixture empirical Bayes and deconvolution across a thousand resamples is impractical. The uncertainty arising from the choice of estimator is therefore carried separately by the systematic bracket rather than by the bootstrap, and the two sources are reported side by side, so that sampling uncertainty and estimator-selection uncertainty are never conflated. The calibration of these intervals is verified by a coverage simulation on synthetic data with a known ground truth. As shown in Appendix~\ref{app:coverage}, the plug-in and NSB ends respectively under-cover and over-cover by design, whereas the bias-corrected mixture empirical Bayes gives the best coverage and the systematic bracket captures the true irreducible error on every dataset.

The robustness of the conservative reachability to these choices is quantified directly. When an outer bootstrap of $200$ resamples re-estimates the noise rate through a cluster bootstrap on the A-side instances in addition to resampling the evaluation instances, the full uncertainty interval of the conservative reachability at the default decay coefficient $\beta=0.7$ is $[0.367, 0.401]$ at L2 and $[0.366, 0.401]$ at L1, essentially the same width as the instance-only interval. This is because the conservative end depends only on the noise rate and not on the prior fit, so prior-fitting uncertainty does not propagate. Across the nine combinations of the decay coefficient $\beta\in\{0.5,0.7,0.9\}$ and three split seeds, the conservative numerator has mean $0.147$ and standard deviation $0.010$ at L2 and mean $0.131$ and standard deviation $0.009$ at L1. A smaller $\beta$ removes more annotation noise and hence lowers the ceiling, so the conservative reachability itself spans $[0.345, 0.411]$ at L2 and $[0.338, 0.409]$ at L1, with a minimum of $0.3375$ at L1 with $\beta=0.5$, which is the value underpinning the headline lower bound of at least about $33$\% across the smoothing-and-seed grid. Finally, the NSB upper end is not a grid artifact. Across integration grids of $120$, $500$, and $2000$ nodes the ceiling drifts by at most $0.0016$ at L2 and $0.0014$ at L1, below the Monte Carlo noise, so the upper-end reachability near $1.03$ is genuine.

Finally, we delineate the statistical framing honestly. The comparisons in this work are estimation rather than between-group significance testing, and the reported intervals are per-quantity instance-bootstrap $95$\% confidence intervals. The cross-label and cross-task comparisons are descriptive and not significance claims from multiple testing, so they are not subject to multiple-comparison correction. The effect size is the reported absolute magnitude itself, namely the bracket width, the spread of the reachability $R$ from $0.38$ to $1.03$, and the ceiling gap between the plug-in and corrected estimators, together with their attendant confidence intervals, and not a standardized test statistic.

For reproducibility, classifier inference is run on an RTX PRO 6000 Blackwell 96GB (transformers 4.44.2, torch 2.2) and ceiling estimation is run on CPU. The fixed dependencies (\texttt{requirements.txt}) and the full code are released, so all numbers are reproducible from the frozen split (seed $20260723$).

\section{Confidence-Interval Coverage Simulation}
\label{app:coverage}

Whether the bootstrap confidence intervals of Appendix~\ref{app:boot} attain nominal coverage is verified directly on synthetic data with a known ground truth. Label distributions $p_i$ are drawn from a known generative distribution $G$, and its dataset-level irreducible error $C_{\mathrm{true}}=\mathbb{E}_G[1-\max_y p_{iy}]$ is fixed by large-scale Monte Carlo. $G$ is a $K=7$ anchored peak-and-splitting mixture with a neutral-dominant base measure that mimics the skew of the GoEmotions L2 marginal, and it belongs to the same distributional family assumed by the mixture empirical Bayes. With $2{,}000{,}000$ prior draws we obtain $C_{\mathrm{true}}=0.261$ with a Monte Carlo standard error of $1.7\times 10^{-4}$. Each synthetic dataset consists of $2{,}000$ instances, and each instance is given $n\in\{3,5\}$ finite annotations by a multinomial, consistent with the mean annotator count of $3.4$ in GoEmotions. For each estimator we compute the same instance-level estimator as in Appendix~\ref{app:mixeb} and the same $B=1000$ instance-bootstrap $95$\% confidence interval as in Appendix~\ref{app:boot}, and across $M_{\mathrm{sim}}=100$ independent synthetic datasets we measure the empirical coverage as the fraction of confidence intervals that contain $C_{\mathrm{true}}$. $M_{\mathrm{sim}}=100$ was chosen from the synchronous CPU-only run time, and the standard error of the coverage near nominal is about $0.022$. The frozen seed is $20260729$.

Table~\ref{tab:coverage} shows the results. The plug-in coverage is $0$ at both $n=3$ and $n=5$, because the plug-in is a negatively biased lower bound owing to the finite-sample upward bias of $1-\max_y \hat{p}$, so its confidence interval lies entirely below $C_{\mathrm{true}}$ on every dataset, with the bias shrinking with the sample size from $-0.072$ at $n=3$ to $-0.044$ at $n=5$. The NSB coverage is also $0$, but because NSB is designed as an upper bound with positive bias $+0.064$ and $+0.040$, its confidence interval lies entirely above $C_{\mathrm{true}}$ on every dataset. The systematic bracket $[\text{plug-in},\ \text{NSB}]$ spanned by the two ends therefore contains $C_{\mathrm{true}}$ on all $100$ datasets, consistent with the framing of this work in which it is the bracket rather than a point estimator that captures the truth. The bias-corrected mixture empirical Bayes is nearly unbiased, with bias of only $+0.001$ and $+0.003$, so it attains the best coverage of the three, reaching $0.75$ at $n=3$ and $0.89$ at $n=5$ and approaching the nominal $0.95$ with the annotation count. The residual under-coverage below nominal nonetheless stems from finite-sample residual bias and slight prior misspecification, and it supports the design choice of reporting the systematic bracket rather than relying on the confidence interval of a single point estimator alone.

\begin{table}[t]
  \centering
  \small
  \setlength{\tabcolsep}{4pt}
  \caption{Empirical coverage of the $95$\% bootstrap confidence intervals. Ground truth $C_{\mathrm{true}}=0.261$, $M_{\mathrm{sim}}=100$ synthetic datasets, $B=1000$. Bold marks the best coverage at each $n$.}
  \label{tab:coverage}
  \begin{tabular}{llccc}
    \toprule
    Estimator & $n$ & Coverage & CI width & Bias \\
    \midrule
    plug-in     & 3 & 0.00 & 0.028 & $-0.072$ \\
    mixture EB  & 3 & \textbf{0.75} & 0.020 & $+0.001$ \\
    NSB         & 3 & 0.00 & 0.023 & $+0.064$ \\
    \midrule
    plug-in     & 5 & 0.00 & 0.027 & $-0.044$ \\
    mixture EB  & 5 & \textbf{0.89} & 0.024 & $+0.003$ \\
    NSB         & 5 & 0.00 & 0.027 & $+0.040$ \\
    \bottomrule
  \end{tabular}
\end{table}

\section{Annotation-Noise Deconvolution: EM and Recovery}
\label{app:deconv}

The mixture model of Eq.~\eqref{eq:deconv} is fitted by EM \citep{gordon2021}. The E step is implemented as a leave-one-out prediction step, because in-sample posterior-mean imputation degenerates at $n_i=3$, as each observation explains itself and $\varepsilon$ collapses to zero. A slightly smoothed leave-one-out prediction with decay coefficient $\beta=0.7$ recovers the identifiability of $\varepsilon$. The update of $\tilde{p}$ is a regularized posterior-mean imputation rather than the maximum a posteriori value, and it therefore has no strict monotonicity guarantee, so monotonicity is monitored empirically and damping is introduced upon a violation. The noise rate $\varepsilon_a$ is estimated once at L2 granularity, and the same $\hat{\varepsilon}_a$ together with the corresponding $\nu$ is reused for the other granularities and the binary decomposition, which avoids the inconsistency of $\varepsilon$ taking different values across granularities.

The validity of this approximation is supported by a recovery experiment on synthetic data. At $n=3$ the noise rate is recovered with mean absolute error $0.039$ and correlation $0.93$, and when the true noise is $\varepsilon=0$ no false positives arise, with an estimated mean of $0.015$. Because the removal effect is confounded with the finite-sample downward bias of the plug-in at $n=3$, the isolation of the removal effect is verified at $n=15$, where the absolute entropy bias decreases from $0.270$ for the raw plug-in to $0.112$ after deconvolution. Furthermore, a recovery experiment that draws the signal from a prior fitted to the actual GoEmotions L2 counts and the annotator counts from the actual distribution (mean $3.4$ across all tracks and $2.3$ on the A side) shows that at these observed counts the noise rate is overestimated with mean absolute error of about $0.16$ and a mean of $0.375$ against a true $0.212$, yet the rank correlation stays high between $0.88$ and $0.96$, and the removal effect cannot be separated from the finite-sample bias until $n=15$, where the error drops to $0.018$. In all real-sample settings the post-deconvolution plug-in floor lies below the true Bayes error, for example $0.216$ against $0.373$. This confirms that the post-deconvolution plug-in end is a conservative lower bound and that the reported irreducible fraction is a lower bound rather than a point estimate. Figure~\ref{fig:reach} visualizes how the reachability $R$ resulting from Section~\ref{sec:e2_reach} depends on the choice of estimator and deconvolution, and shows that no single reachability value exists.

\section{Validation of the Estimators on Synthetic Data}
\label{app:tests}

The estimators are validated on synthetic distributions with known entropy. On the uniform distribution with $K=7$ and $n=3$, the entropy bias $\mathbb{E}[\hat{H}]-H_{\mathrm{true}}$ in bits is $-1.498$ for the plug-in, $-1.118$ for Miller--Madow, $-0.789$ for NSB, and $-0.781$ for the symmetric Dirichlet with concentration $0.5$, and on the uniform distribution with $K=28$ and $n=3$ the plug-in bias reaches $-3.298$, confirming a bias exceeding $1$ bit in the severely undersampled regime. The Fano inverse at $K=2$ agrees with $\min(p,1-p)$ to an error of $1.4\times 10^{-14}$, and the closed-form binary Bayes-error floor of Eq.~\eqref{eq:binclosed} agrees with quadrature evaluation to a maximum error of $3.75\times 10^{-12}$.

\section{Irreducible-Error Floors for All Labels}
\label{app:perlabel}

\begin{figure*}[t]
  \centering
  \includegraphics[width=\textwidth]{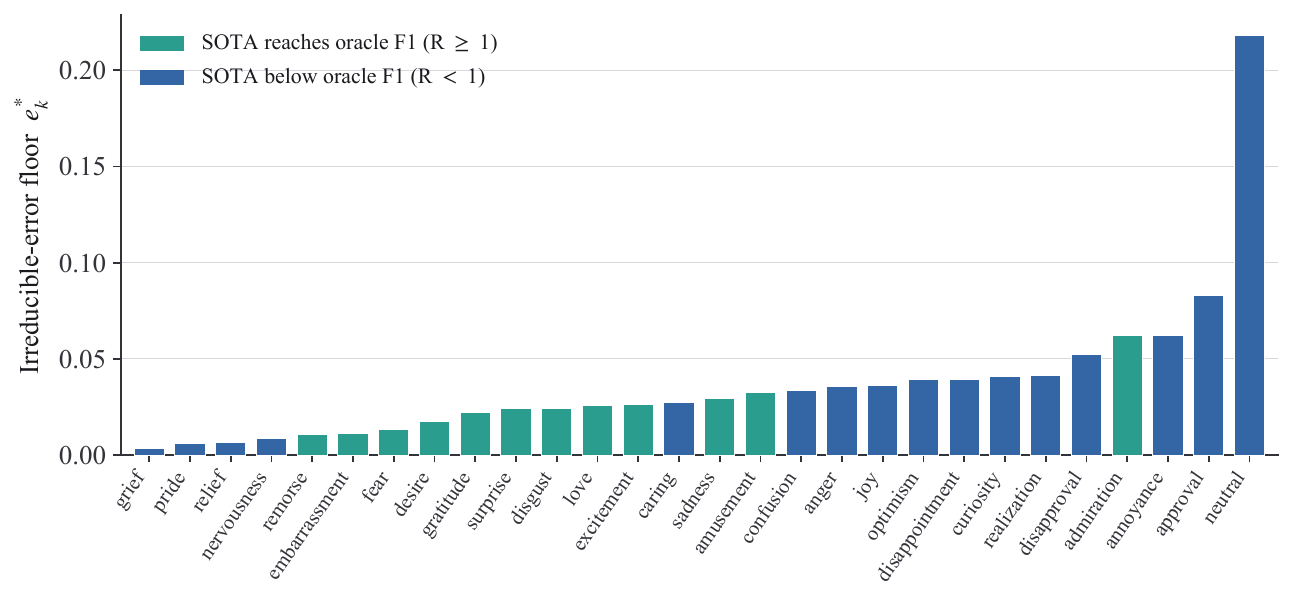}
  \caption{Per-label headroom.}
  \label{fig:headroom}
\end{figure*}

Table~\ref{tab:perlabel} lists the instance-level irreducible-error floor $e^{*}_k$ for all $28$ GoEmotions labels of the binary decomposition, of which the main text in Section~\ref{sec:e1} reports only representative values. The floor is smallest for rare, high-agreement emotions and largest for frequent, ambiguous emotions. Figure~\ref{fig:headroom} shows the per-label headroom of Section~\ref{sec:e2}, where reaching the oracle F1 is unrelated to the height of the floor. For binary F1, the oracle marginal-threshold floor is $0.357$ at macro and $0.407$ at micro, whereas the evaluated classifier at threshold $0.5$ attains $0.333$ at macro and $0.438$ at micro, with the micro reachability exceeding $1$ because the floor is a marginal reference value. The soft-label P3 reachability against the cross-entropy floor spans $0.28$ to $0.85$ at L2 and $0.30$ to $0.86$ at L1. As a diagnostic, $\hat{G}$ takes negative values from $-0.001$ to $-0.013$ at the upper L2 ceiling, namely the raw NSB counts and the post-deconvolution mixture empirical Bayes and NSB, and this is reported as an alarm rather than as superhuman performance, whereas at the conservative end all gaps are positive.

\begin{table*}[t]
  \centering
  \small
  \setlength{\tabcolsep}{10pt}
  \caption{Irreducible-error floors $e^{*}_k$ for all $28$ labels (ascending in the floor).}
  \label{tab:perlabel}
  \begin{tabular}{lclclclc}
    \toprule
    Label & $e^{*}_k$ & Label & $e^{*}_k$ & Label & $e^{*}_k$ & Label & $e^{*}_k$ \\
    \midrule
    grief         & 0.0033 & desire      & 0.0175 & sadness      & 0.0296 & curiosity   & 0.0407 \\
    pride         & 0.0063 & gratitude   & 0.0220 & amusement    & 0.0326 & realization & 0.0415 \\
    relief        & 0.0063 & surprise    & 0.0242 & confusion    & 0.0336 & disapproval & 0.0523 \\
    nervousness   & 0.0087 & disgust     & 0.0243 & anger        & 0.0358 & admiration  & 0.0620 \\
    remorse       & 0.0106 & love        & 0.0258 & joy          & 0.0359 & annoyance   & 0.0622 \\
    embarrassment & 0.0114 & excitement  & 0.0264 & optimism     & 0.0393 & approval    & 0.0827 \\
    fear          & 0.0133 & caring      & 0.0271 & disappoint.\ & 0.0393 & neutral     & 0.2178 \\
    \bottomrule
  \end{tabular}
\end{table*}

Figure~\ref{fig:map} visualizes the cross-task predictability map of Section~\ref{sec:map} together with the base rates, and shows that without controlling for the base rate the ceilings are not comparable across tasks. For the binary tasks with a base rate of about $0.3$, the irreducible ambiguity is largest for irony at $[0.206, 0.261]$, followed by offensiveness at $[0.173, 0.212]$ and the presence of emotion at $[0.116, 0.149]$, which is interpreted as arising because irony is the most context-dependent and the most dispersed across annotators. Deconvolution lowers the floor on every binary task, from $0.206$ to $0.139$ for irony and from $0.173$ to $0.137$ for offensiveness, so that irony and offensiveness nearly converge. The mean noise rate $\bar{\varepsilon}$ of irony is also the largest at $0.257$, suggesting that a substantial part of the observed disagreement stems from annotation noise. When emotion intensity is treated as an ordinal $K{=}4$, the BRIGHTER floor rises to $[0.173, 0.221]$, confirming that the floor rises monotonically with the number of classes. Because BRIGHTER lacks annotator identifiers, deconvolution and the A/B split cannot be applied, and its multiply annotated data has a narrow bracket width of about $0.05$, confirming that the impact of undersampling is small.

\section{Cross-Task SOTA Decomposition}
\label{app:crosstask}

Table~\ref{tab:crosstask} decomposes off-the-shelf public classifiers on each cross-task corpus against their conservative post-deconvolution plug-in ceiling, extending the reachability analysis of Section~\ref{sec:e2_reach} beyond GoEmotions. The offensiveness classifier is cardiffnlp/twitter-roberta-base-offensive and the irony classifier is cardiffnlp/twitter-roberta-base-irony, each thresholded to the binary positive class. The emotion row is SamLowe/roberta-base-go\_emotions (macro across six emotions), mapped and averaged onto the six BRIGHTER emotions through a noisy-OR over the GoEmotions labels.

Two observations transfer from GoEmotions. First, the conservative irreducible component remains strictly positive on all three corpora, at $0.137$ for offensiveness, $0.139$ for irony, and $0.116$ for emotion, so a positive irreducible error $C>0$ is not a GoEmotions artifact. Second, the reachability again depends strongly on the estimator. The conservative post-deconvolution plug-in end is the smallest on every task, and the raw empirical Bayes and NSB ends raise it by roughly $1.3$ to $1.9$ times, from $R_{\mathrm{cons}}=0.413$ to $R_{\mathrm{EB}}=0.640$ for offensiveness, from $0.359$ to $0.672$ for irony, and from $0.422$ to $R_{\mathrm{NSB}}=0.543$ for emotion, so the estimator dependence of the saturation verdict generalizes across tasks.

These reachabilities are lower bounds and are honestly scoped in two respects. The classifiers are public models that are not fine-tuned on the target corpora, so their error is inflated relative to an in-domain model, and the reported $R$ therefore underestimates the reachability that a fine-tuned classifier would attain. Accordingly, the cross-task $R_{\mathrm{cons}}$ of $0.36$ to $0.42$ is comparable to the conservative band of $0.29$ to $0.39$ for GoEmotions, whereas the raw ends of $0.48$ to $0.67$ fall below the nearly unit in-domain reachability of GoEmotions in Section~\ref{sec:e2}. The emotion row further involves a domain and label mismatch. The six BRIGHTER emotions are matched through a synthetic mapping in which categories such as fear--anxiety and social-warmth are composites of several GoEmotions labels, which lowers their reachability further. The per-emotion reachabilities are not cherry-picked and span from $0.297$ for the composite fear--anxiety category to $0.540$ for anger, with both ends reported. The cross-task decomposition therefore supports the generality of the two pillars as a qualitative phenomenon rather than as a transfer of the GoEmotions numbers.

\end{document}